# An LLM-Orchestrated Multi-Modal Chatbot for Tomato Disease Consultant

Ngoc-Son Le[1], Khanh-Nghi Phan[2] and Ngoc Hong Tran[2]*

## Abstract

This paper introduces a chatbot developed to assist in diagnosing common diseases in tomato plants through natural language interaction. Users can describe symptoms in text. Then, the chatbot responds with the most likely disease, and basic treatment suggestions and related image-based reasoning. The system leverages pre-trained Mistral and Llama 3.2 large language models. We also use LangChain and Streamlit to provide a lightweight and offline-friendly interface. This work targets to support tomato growers a fast and accessible tomato plant health guidance when the technical expertise is not available nearby, especially in the remote areas. The experimental results are promising. Llama 3.2 achieved higher semantic relevance, while Mistral provided faster and more lightweight responses in several test cases.

**Keywords:**

AI Assistant, Large Language Model (LLM), Retrieval-Augmented Generation (RAG), Tomato Plant Disease.

## Introduction

Global agricultural productivity is seriously threatened by plant diseases, particularly in widely grown crops like tomatoes. Tomato is one of vegetables extremely vulnerable to a variety of pathogens, pests, and physiological conditions (Savary et al., 2019; Panno et al., 2021). Expert inspection and laboratory testing are key components of traditional disease identification methods. However, they are frequently expensive, time-consuming, and unavailable to smallholder farmers.

Previous studies have demonstrated the effectiveness of deep learning techniques for plant disease detection and classification (Boulent et al., 2019). Mohanty et al. (2016) applied convolutional neural networks (CNNs) for image-based plant disease recognition and achieved high classification accuracy across multiple crop species. Moreover, Fuentes et al. (2018) proposed a high-performance deep neural network framework for detecting tomato diseases and pests. They used refinement filter banks to improve recognition performance under complex conditions. In another study, Fuentes et al. (2017) developed a robust real-time deep learning-based detector. This detector is capable of identifying tomato plant diseases and pests directly in field environments with high detection accuracy and efficiency. However, it is still a challenge for consulting the farmer to solve the matter they found out from the images.

[1] Faculty of Information Technology, University of Science, Vietnam National University, Ho Chi Minh City, 700000, Vietnam

[2] Computer Science Program, Faculty of Engineering, Vietnamese-German University, Ho Chi Minh City, 700000, Vietnam

*Corresponding Author: ngoc.th@vgu.edu.vn

In order to support the tomato farmers and gardeners at any time, we introduce a chatbot system. We uses natural language interaction and optional image based inputs to help users diagnose common tomato plant diseases. The chatbot system supports quick, easily accessible, and contextually relevant responses for early tomato disease detection. Besides, the chatbot also provides plant care recommendations by combining machine learning-based classification with semantic retrieval.

More particularly, this work integrates Large Language Models (LLMs) (Zhao et al., 2023) with a Retrieval-Augmented Generation (RAG) framework (Patrick et al., 2020) to provide more accurate and context-aware responses to user inquiries. The chatbot retrieves information from a domain-specific vector database. This database was constructed from a manually curated dataset of tomato diseases. It contains structured metadata and related disease images collected from academic publications and trusted agricultural institutions (Pitchayagan et al., 2021; PlantVillage, 2018; University of Minnesota Extension, 2024). In addition, the chatbot system also displays relevant disease images corresponding to the detected condition to improve user understanding and diagnosis support. Furthermore, the integration of MXBAI-Embed-Large (Sean et al., 2024) (that is, the embedding model and Mistral (Li & Li, 2023)) deployed locally through open source Ollama framework, enables efficient offline-capable operation and ensures reliable functionality in low-connectivity agricultural environments (Jiang et al., 2024).

The remainder of this paper is organized as follows. Section 2 presents the research objectives. Section 3 presents the literature reviews. The research methodology including the proposed chatbot framework, data source, and functionality workflow is presented in Section 4. Section 5 demonstrates the results including evaluation metrics and result findings. Eventually, discussion and conclusion are presented in Section 6.

## Literature Review

Recent advances in artificial intelligence have significantly improved the development of intelligent agricultural support systems, particularly for plant disease diagnosis. Earlier studies on image-based disease classification used convolutional neural networks (CNNs), where tomato leaf datasets were used to train models capable of recognizing visual disease patterns with high accuracy (Mohanty et al., 2016). Moreover, the deep learning models, such as ResNet, MobileNet, and EfficientNet, are also used to identify tomato leaf diseases under controlled conditions (Too et al., 2019). However, they are mainly image classification. So, they limited interaction capabilities for non-technical users in real farming environments.

Furthermore, the conversational AI and Large Language Models (LLMs) have been used in agricultural decision support systems. However, they typically depend on rule-based responses or predefined templates. Therefore, they limit flexibility when they handle diverse user queries (Pitchayagan et al., 2021; Suebsombut et al., 2023). The emergence of LLMs has enabled conversational systems to understand natural language instructions and generate context-aware recommendations (Brown et al., 2020). Zhao et al. (2024) integrated LLMs with agricultural knowledge graphs to improve plant disease detection and agricultural reasoning capabilities. Similarly, AgroLLM (Ravindran et al., 2026) employed Retrieval-Augmented Generation (RAG) with large language models to provide agricultural knowledge assistance with improved contextual accuracy. In another study, researchers integrated AI-based pest detection with language models to support real-time tomato cultivation management (Sahin et al., 2025). Furthermore, Chat Demeter (Zhang, 2026) demonstrated the potential of multi-agent conversational systems integrating CNN-transformer architectures for plant disease diagnosis. Emerging agricultural multimodal large language models such as AgriChat (Boudiaf et al., 2026) further indicate that combining visual reasoning and conversational intelligence is becoming a major research direction in smart agriculture.

Under above motivation, in this work, we deployed LLM based Chatbot combining visual reasoning to support the farmers and gardeners easier to comprehend the matter.

# Methodology

The chatbot system combines several modern Python libraries and tools to implement a lightweight, locally-deployable diagnostic assistant. The main components include LangChain, Ollama, Chroma, Streamlit, and pre-trained models like Mistral and MXBAI-Embed-Large. Each plays a unique role in the Retrieval-Augmented Generation (RAG) architecture and is described in detail below.

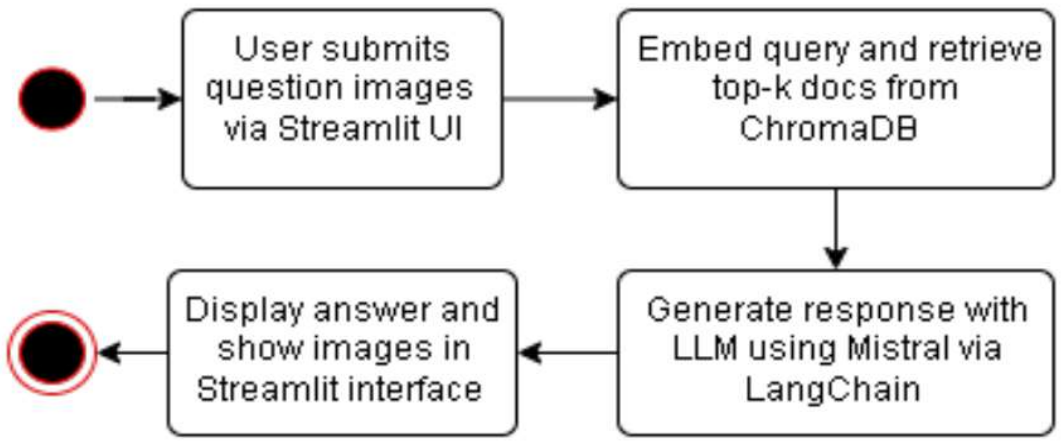

**Figure 1.** Action-based pipeline for tomato disease diagnosis and visual output

## Pipeline Flow

The system processes each user query through a four-stage pipeline (see Figure 1). First, the system transforms natural language input into enriched diagnostic feedback with visual support. The process start executing when the user submits a question through the Streamlit interface. This input typically includes a natural language description of symptoms observed on tomato plants. Then, LangChain transforms the query into a vector by using the MXBAI-Embed-Large embedding model. This vector searches ChromaDB for the top $k$ most semantically similar disease documents. The retrieved documents and user query are sent to the Mistral language model in LangChain's prompt template. The LLM generates a detailed answer with diagnosis, explanation, and treatment suggestions. Then, the response is shown in the Streamlit UI. If the response contains recognized disease labels, the system obtains the related images in the dataset. Finally, the UI displays them with the text for visual clarity and user validation.

# Chatbot Architecture

Figure 2 describes the architecture of the tomato disease diagnosis chatbot system. The application is organized into 4 tiers as follows.

*User Tier:* is the entry point. The users submit questions and receive responses at this tier.

*Front-end Tier:* uses Streamlit to handle the user interface. This tier captures user input, then it sends queries to the backend. After that, the textual and image-based results are displayed. The front end tier can also show chat history.

*Back-end Tier:* involves the LangChain orchestrator. This tier combines prompt templates. Then, it can retrieve relevant documents. The back end tier can also manage the flow of information between the frontend and the data tier.

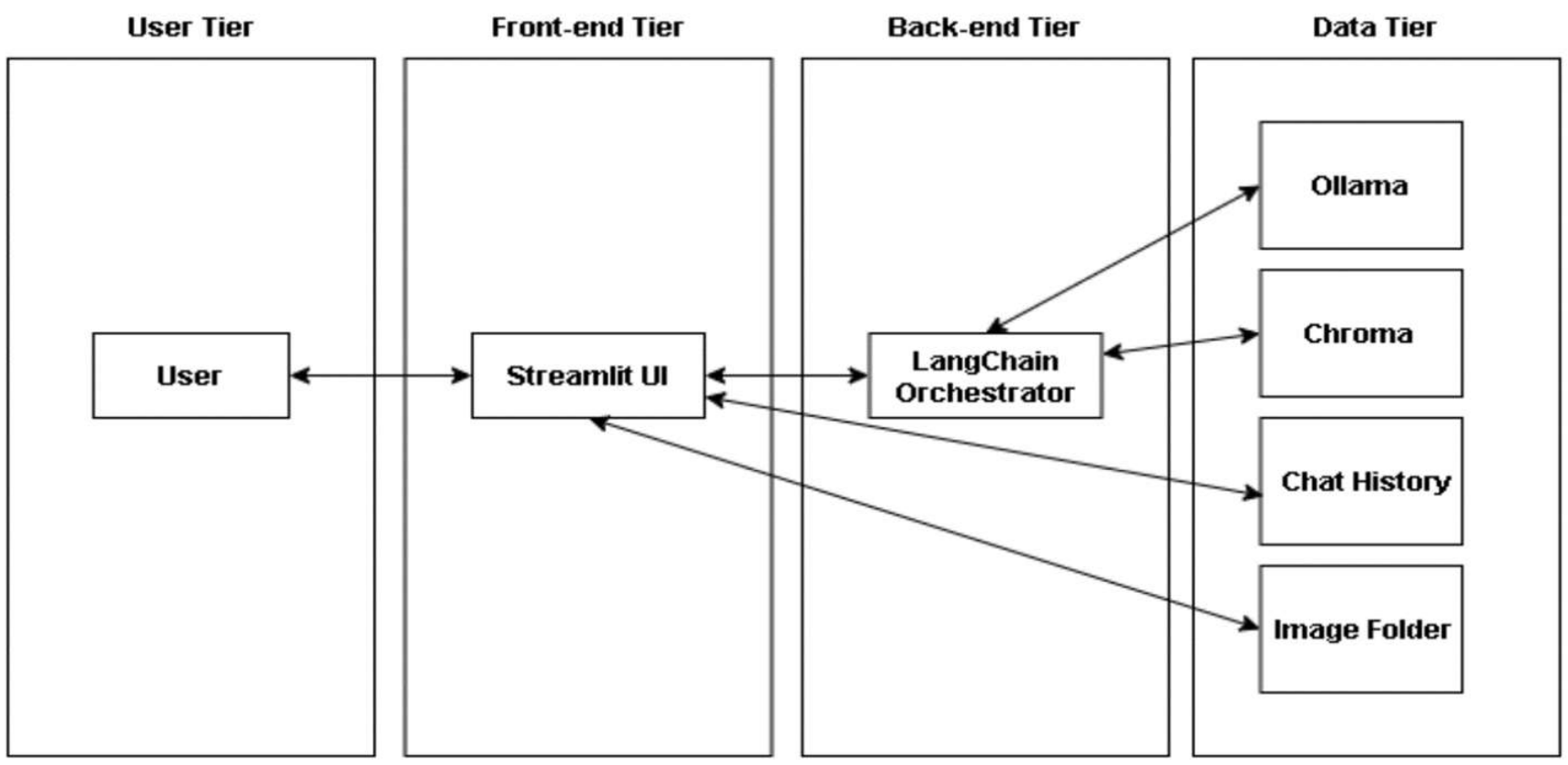

**Figure 2.** System architecture and component interaction in the tomato disease Chatbot

*Data Tier:* In this tier, Ollama provides LLM inference capabilities by local models. ChromaDB performs semantic retrieval for top-k relevant documents based on the vector similarity. Chat history stores past conversations in structured JSON format for continuity and offline access. Image folder: contains labeled images of tomato diseases. They are fetched and displayed when a disease label is detected in the LLM's response.

In this architecture, the users input their queries in the UI. Then, the UI passes the queries to LangChain. On its turn, LangChain retrieves relevant documents from ChromaDB. Then, it sends the prompt to Ollama for completion. Next, the response is parsed. After that, any detected disease labels are used to fetch corresponding images from the image folder. The final response and images are shown to the user. Finally, the interaction is saved into the chat history.

This architecture is chosen for its ability to balance

- Offline capability as all components run locally, therefore, it supports deployment in low connectivity environments.
- Latency as local inference with Mistral and fast vector search with Chroma ensure low response times.
- Modularity as components can be upgraded independently (e.g., swapping MXBAI for another embedding model or replacing Streamlit with a mobile UI). This modular and efficient design contributes to the system's usability in critical agricultural field conditions, such as real-time, interpretable, and visual feedback.

## Data Source

This dataset was manually collected from the academic and agricultural sources. This custom dataset focuses on tomato plant diseases, pests, and disorders. It supports for both the visual and retrieval based components of the system. In particular, there are two core parts in the dataset.

## Text Dataset

This dataset contains 45 entries in JSON format. Each entry is a tomato-related disease, pest, or disorder. The top data sample is presented in Figure 3 in .csv format before it is converted to JSON format as in Figure 4.

| disease_name | description | symptoms | treatment | pesticides_or_medicine | label |
|---|---|---|---|---|---|
| Tomato Mosaic virus | Tomato mosaic virus (ToMV for short) is an infectious disease of tomatoes and one of the most widespread viruses affecting this vegetable, of which various strains are known. | Damage most commonly occurs on the leaves, which show mottled light green and dark green spots – rather like a mosaic effect – and deformations such as bubbling and crinkling. "Other strains also alter the appearance of the leaves, while still others affect only the fruits – i.e. the tomatoes – or the entire plant (stem, leaves and fruits)." | Eliminate diseased plants immediately and dispose of them well away from the vegetable patch: there must be no leftover remains, including in the soil.Disinfect the tools you use on the vegetable patch with bleach – your pruning shears, for example – and wash your hands with soap and water before handling the plants. | There are no targeted treatments against tomato viral diseases including ToMV, prevention is the best way to combat them. | MosaicVirus |
| Tomato Yellow Leaf Curl Virus | Tomato yellow leaf curl virus (TYLCV) can infect over 30 different kinds of plants, but it is mainly known to cause devastating losses of up to 100 per cent in the yield of tomatoes. Both field and glasshouse grown tomatoes are susceptible. TYLCV is vectored by silverleaf whitefly (Bemisia tabaci).<br><br>TYLCV is a begomovirus, implying it has a wide host range on dicotyledons. There are two strains of TYLCV, a mild strain and a severe strain. | The most obvious symptoms in tomato plants are small leaves that become yellow between the veins. The leaves also curl upwards and towards the middle of the leaf. | The TYLCV strain occurring in Australia is not known to be seed transmitted or to be transmitted mechanically.<br><br>TYLCV infected plants tend to be found randomly or in patches in a crop.<br><br>The world-wide spread of TYLCV is suspected to be associated with the movement of commercial tomato and ornamental plants contaminated with virus infected whitefly. There are no treatments for viruses. Infected cucurbits plants cannot be cured; therefore, the focus should be on prevention. | TYLCV is mainly spread by whiteflies (Bemisia tabaci). The TYLCV strain occurring in Australia is not known to be seed transmitted or to be transmitted mechanically.<br><br>TYLCV infected plants tend to be found randomly or in patches in a crop.<br><br>The world-wide spread of TYLCV is suspected to be associated with the movement of commercial tomato and ornamental plants contaminated with virus infected whitefly. | YellowCurl |
| Septoria leaf spot | Septoria leaf spot is caused by a fungus, Septoria lycopersici. It is one of the most destructive diseases of tomato foliage and is particularly severe in areas where wet, humid weather persists for extended periods.<br>Septoria leaf spot is one of two common fungal diseases that can devastate tomatoes in both commercial settings and home gardens. | Septoria leaf spot usually appears on the lower leaves after the first fruit sets. Spots are circular, about 1/16 to 1/4 inch in diameter with dark brown margins and tan to gray centers with small black fruiting structures. Characteristically, there are many spots per leaf. This disease spreads upwards from oldest to youngest growth. If leaf lesions are numerous, the leaves turn slightly yellow, then brown, and then wither. Fruit infection is rare. Symptoms of Septoria leaf spot first appear at the base of affected plants, where small (approximately ¼ inch diameter) spots appear on leaves and stems. These spots typically have a whitish center and a dark border. Eventually multiple spots on a single leaf will merge, leading to extensive destruction of leaf tissue. Septoria leaf spot can lead to total defoliation of lower leaves and even the death of an infected plant. | Once symptoms of Septoria leaf spot appear, control is difficult. Thinning of whole plants or removal of selected branches from individual plants may slow the disease by increasing airflow and thus reducing humidity and the length of time that leaves remain wet. Septoria leaf spot is best controlled using preventative measures. Destroy infested plants by burning (where allowed by local ordinance) or burying them. Rotate vegetables to different parts of your garden each year to avoid areas where infested debris (and thus spores of Septoria lycopersici) may be present. Use Septoria leaf spot-resistant tomato varieties whenever possible. Increase spacing between plants to increase airflow and decrease humidity and foliage drying time.<br>Mulch your garden with approximately one inch of a high quality mulch, but DO NOT overmulch as this can lead to wet soils that can contribute to increased humidity. | Fungicides labeled for use on vegetables and containing copper or chlorothalonil may also provide control of Septoria leaf spot if they are carefully applied very early in the course of the disease (before symptoms develop is best) and on a regular basis throughout the rest of the growing season.<br>Apply chlorothalonil, mancozeb, or a copper-based fungicide, such as copper hydroxide, copper sulfate, or copper oxychloride sulfate. Follow harvest restrictions listed on the pesticide label. | Septoria |
| Spider mites Two spotted spider mite | The two-spotted spider mite is the most common mite species that attacks vegetable and fruit crops in New England. Spider mites can occur in tomato, eggplant, potato, vine crops such as melons, cucumbers, and other crops. Two-spotted spider mites are one of the most important pests of eggplant. | Adult TSSM feed by sucking chlorophyll out of the leaves, creating blotchy pale to reddish-brown spots. Feeding injury often gives the top leaf surfaces a mottled or speckled, dull appearance. Leaves then turn yellow and drop. Large populations produce visible webbing that can completely cover the leaves."<br>Other symptoms caused by either severe or constant attack include distorted leaves, overall loss of plant vigor (in spite of adequate moisture and nutrition), whitening or spotting of leaves, yellowing of the plant or some of the leaves, and in some cases loss of foliage and death. | Overhead irrigation or prolonged periods of rain can help reduce populations.<br>Preventative releases of the predatory mite, Phytoseiulus persimilis, may suppress TSSM populations in greenhouses and vegetable fields. Avoid early season, broad-spectrum insecticide applications for other pests.<br>Do not over-fertilize. Outbreaks may be worsened by excess nitrogen fertilization. | bifenazate (Acramite): Group UN, a long residual nerve poison<br>abamectin (Agri-Mek): Group 6, derived from a soil bacterium<br>spirotetramat (Movento): Group 23, mainly affects immature stages<br>spiromesifen (Oberon 2SC): Group 23, mainly affects immature stages<br>OMRI-listed products include: insecticidal soap (M-Pede), neem oil (Trilogy), soybean oil (Golden Pest Spray Oil) | SpiderMite |
| | | Symptoms begin on older leaves as pale green to yellow spots on | Reduce humidity by increasing air circulation through spacing | Fungicides should be applied before infection under favorable | |

**Figure 3.** Preview of the disease dataset as structured in Excel before JSON conversion

```
{
  "No.": 1,
  "disease_name": "Tomato Mosaic virus",
  "description": "Tomato mosaic virus (ToMV for short) is an infectious disease
    of tomatoes and one of the most widespread viruses affecting this vegetable,
    of which various strains are known.",
  "symptoms": "Damage most commonly occurs on the leaves, which show mottled
    light green and dark green spots - rather like a mosaic effect - and
    deformations such as bubbling and crinkling. Other strains also alter the
    appearance of the leaves, while still others affect only the fruits - i.e.
    the tomatoes - or the entire plant (stem, leaves and fruits).",
  "treatment": "Eliminate diseased plants immediately and dispose of them well
    away from the vegetable patch: there must be no leftover remains, including
    in the soil. Disinfect the tools you use on the vegetable patch with bleach
    - your pruning shears, for example - and wash your hands with soap and water
    before handling the plants.",
  "pesticides_or_medicine": "There are no targeted treatments against tomato
    viral diseases including ToMV, prevention is the best way to combat them.",
  "label": "MosaicVirus"
}
```

**Figure 4.** Sample JSON entry from text dataset

Each data point includes the following fields.

- *Disease name* of the disease or disorder.
- *Description*: A scientific summary of its cause and nature.
- *Symptoms:* Typical visual indicators to aid diagnosis.
- *Recommended treatment methods*: Cultural and chemical recommendations.
- *Approved pesticides or remedies*: Specific agents (where applicable).
- *A unique label (used for classification and image linkage):* A unique identifier used to link to image folders. In this work, we collected 45 tomato leaf diseases class, that is, *MosaicVirus, YellowCurl, Septoria, SpiderMite, LeafMold, LateBlight, EarlyBlight, BacterialSpot, Anthracnose, BlackMold, FusariumWilt, BotrytisBlight, TargetSpot, VerticilliumWilt, BacterialCanker, BacterialSpeck, Bacterial Wilt, BlossomEndRot, Catface, Edema, LittleLeaf, MagnesiumDeficiency, NitrogenDeficiency,*

*FertilizerBurn, PhosphoruDeficiency, PotassiumDeficiency, Sunscald, WaterStress, Zippering, BuckeyeRot, SpottedWilt, PeachAphid, BeetArmyworm, ColoradoPotatoBeetle, Cutworms , FleaBeetles, Hornworms, LeafMiner, Looper, StinkBugs, Thrips , Fruitworm, CurlyTop, RootKnot, SpiderMites*.

This file is processed in Python using the LangChain and Chroma libraries to generate a vector store. The embedded documents enable the chatbot to semantically retrieve relevant content during user interactions.

## Image dataset

A directory named TomatoDataset stores 45 subfolders labeled by disease (see Figure 5).

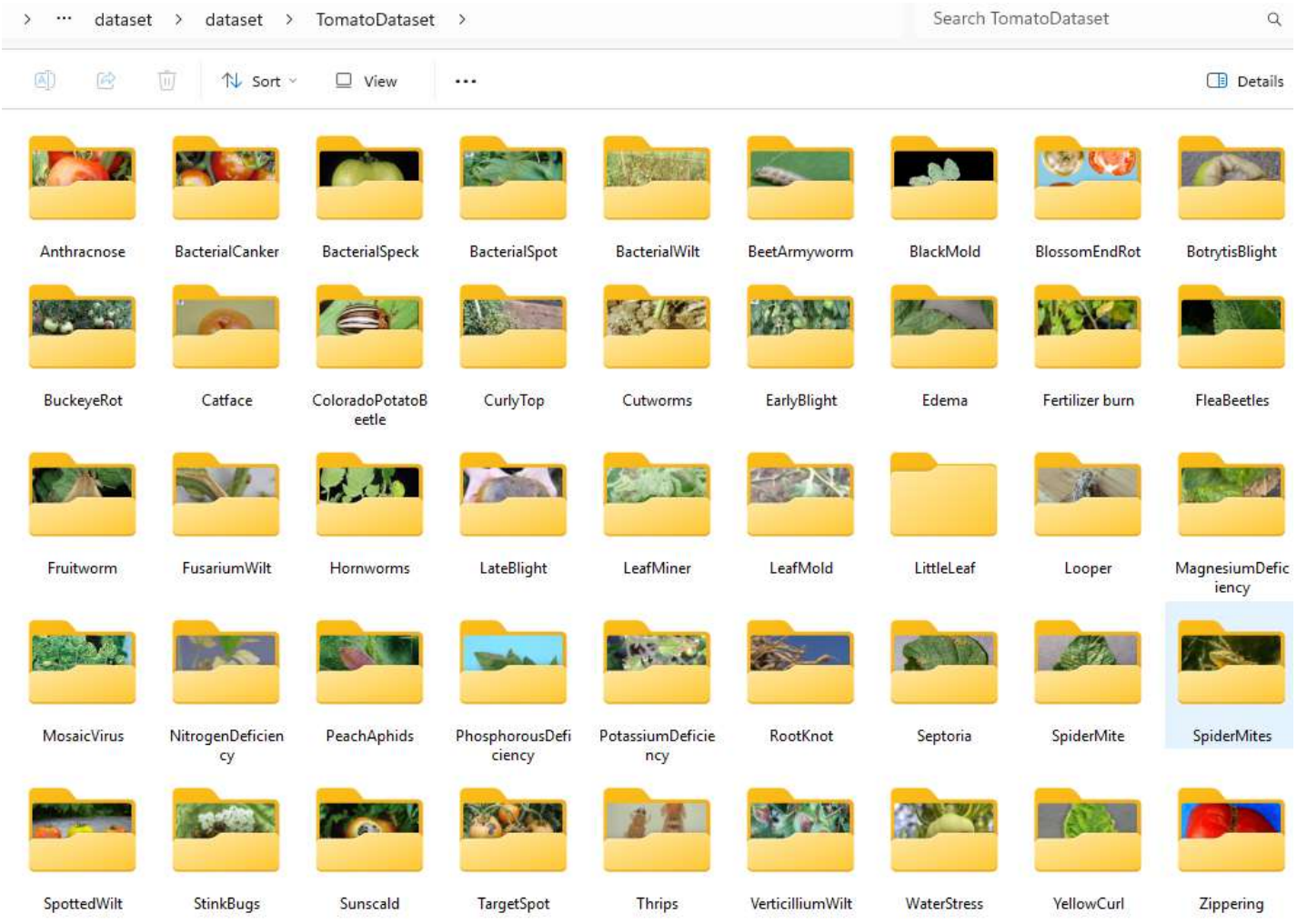

**Figure 5.** Image dataset of 45 tomato leaf disease types

Each subfolder contains five representative images of the corresponding disease or pest, collected manually from trusted sources. These images are automatically shown in the chatbot interface when a disease is detected or inferred. Some samples of tomato disease are depicted in Figure 6 and Figure 7.

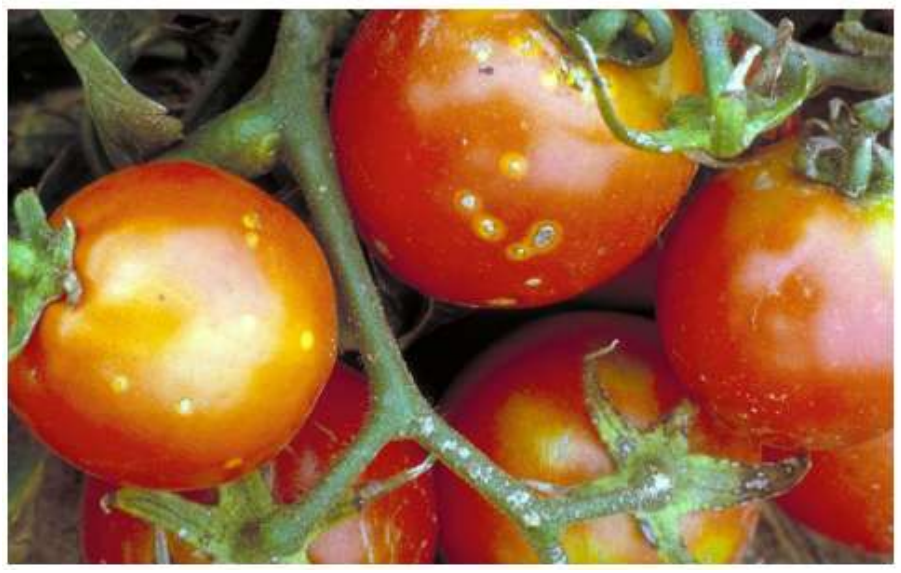

**Figure 6.** Bacterial Canker affecting mature red tomatoes (symptoms: bird's-eye spots)

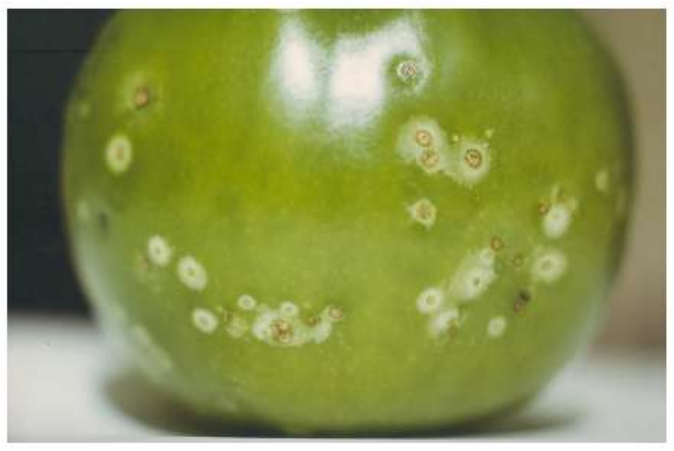

**Figure 7.** Early-stage lesions of Bacterial Canker on green tomato

To ensure accuracy and reliability, Images used in the dataset were collected from agricultural extension resources and publicly available research datasets, including University of Minnesota Extension (University of Minnesota Extension, 2024), University of Florida IFAS (University of Florida IFAS Extension, 2006), PlantVillage (PlantVillage, 2018), NC State Extension (NC State Extension, 2017a; NC State Extension, 2017b), and prior deep learning studies on tomato disease recognition (Mohanty et al., 2016; Fuentes et al., 2018).

# Functionality Workflow

## Chatbot System Implementation

First, we implement semantic indexing with embeddings and vector stores. The vector.py script initializes the semantic search engine by embedding structured tomato disease data and by storing it in a persistent vector database using the Chroma framework. This retriever matches user queries to disease descriptions. This modular design ensures that documents are embedded once and reused efficiently. The combination of Ollama embeddings and Chroma's persistent vector database enables fast, memory-efficient, and semantically meaningful retrievals during chatbot conversations.

Figure 8 visualizes the key steps and function calls that occur when the script is executed. The logic begins with loading the dataset and culminates in building a retriever for semantic search. For loading the dataset, the pipeline begins by reading the structured tomato disease data in CSV format.

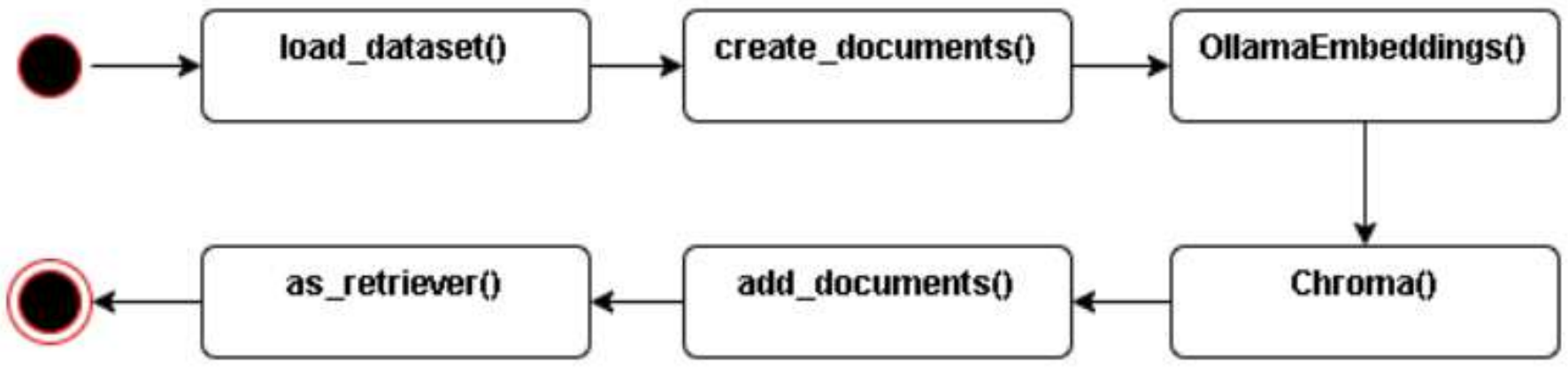

**Figure 8.** Workflow of the vector.py embedding pipeline

Then, to create LangChain documents, each row of the dataset is converted into a LangChain document with structured fields concatenated into a single text block. Metadata is preserved for downstream use. After that, to initialize embedding model, we use OllamaEmbeddings to load the mxbaiembed-large model, which provides strong semantic representation for vector-based search. To initialize Chroma vector store, the Chroma engine is initialized with the embedding function and collection parameters. It will persist vectors to the specified directory. If the vector DB folder does not exist (i.e., the first run), documents are embedded and stored. This avoids reprocessing on subsequent runs. Finally, a retriever is returned, exposing the top-k relevant documents for use in downstream LangChain pipelines. Final Output When the file is loaded, the retriever is initialized immediately and made accessible globally.

## Automating Cloud-Based LLM Hosting and Access Setup

To reduce local hardware requirements while still leveraging GPU-accelerated inference, this project adopts a hybrid architecture. The chatbot's frontend (i.e., Streamlit-based UI) runs locally, while the large language models (i.e., Mistral and MXBAI-Embed-Large) are executed remotely on a Google Colab instance (see Figure 9). This setup allows high-performance model inference via Ollama, and the model is exposed through a public URL using ngrok.

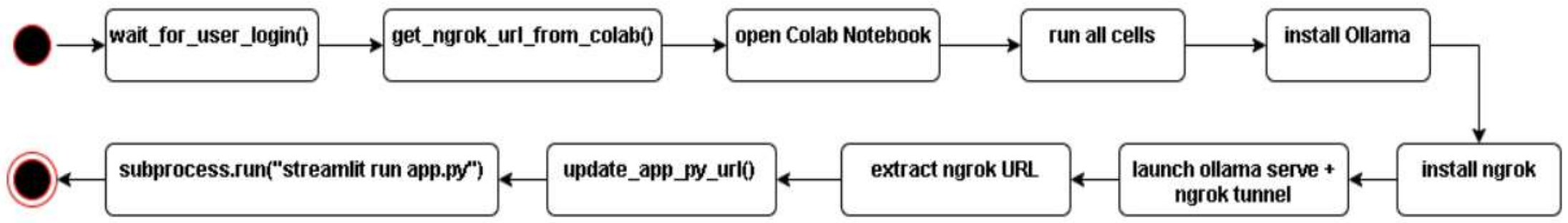

**Figure 9.** Automated remote deployment of Ollama via Google Colab and dynamic reconfiguration in main.py

The orchestration of this remote backend is handled entirely by main.py file, which

- Initiates user authentication with Google Colab.
- Opens and runs a preconfigured Colab notebook to launch Ollama.
- Extracts the ngrok URL which is published by the notebook.
- Dynamically updates the local app.py backend configuration.
- Pulls the required models and launches the chatbot via Streamlit.

## Chat Session Management and Disease Image Retrieval Utilities

The chatbot system incorporates a set of core functions. There are two critical backend tasks: (1) persistent chat session management; (2) image retrieval for diagnosed plant diseases. These functions are encapsulated in the utils.py file. They work as the backbone for chat history continuity and visual support in responses. Figure 10 illustrates the logical sequence of user-driven events and the corresponding backend operations. Each function responds to a specific user action. E.g., it starts a new chat, it views past sessions, it interacts with image-based disease diagnosis.

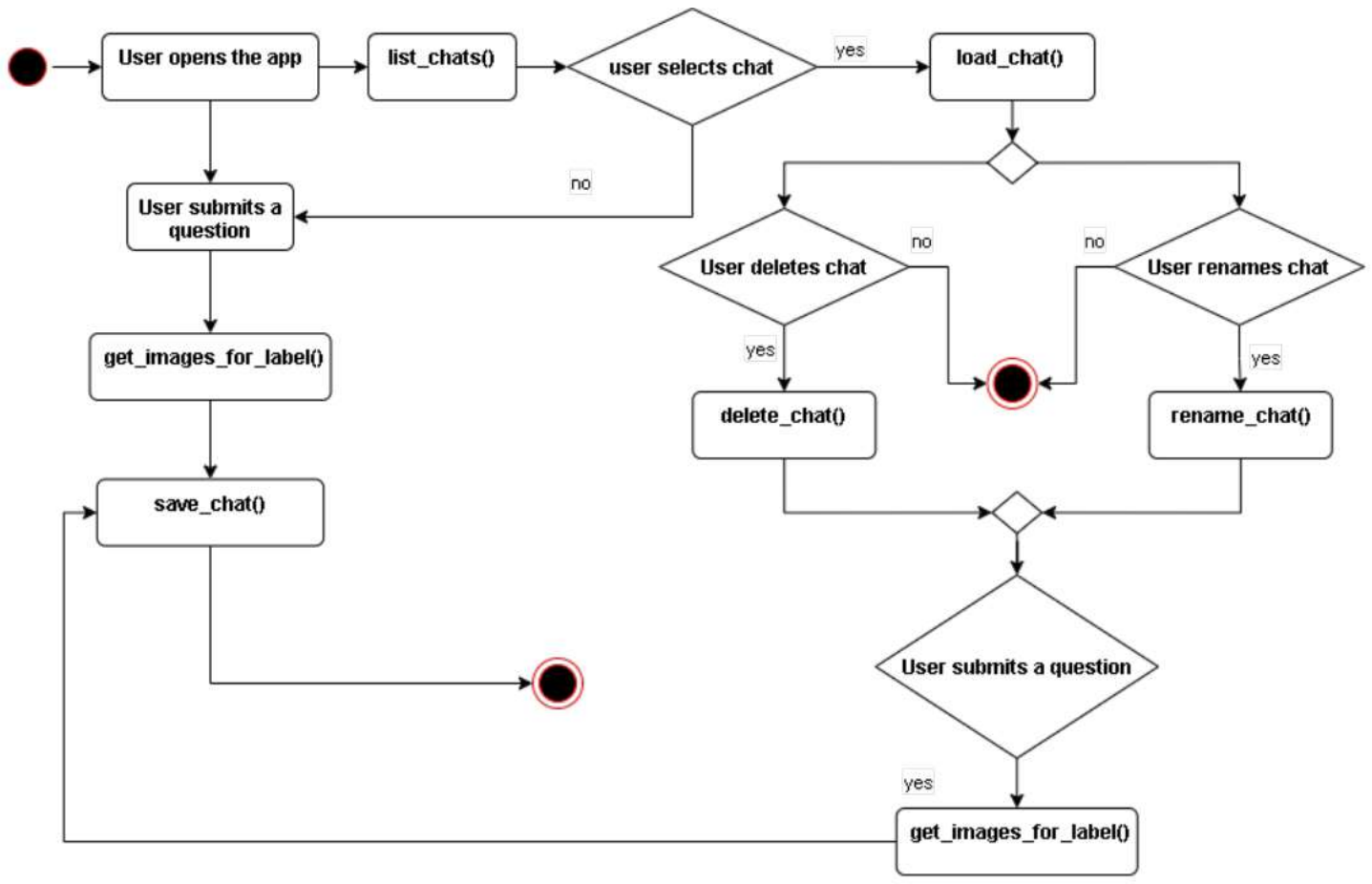

**Figure 10.** Workflow of chat session management and image retrieval logic

## Main Application Logic and Interface

The structure and operation of the Streamlit application displays the Chatbot interface. Besides, it also handles user interaction, invokes the language model, and manages persistent chat history. The logic is implemented in a modular fashion to separate UI handling, language model integration, and data storage. Figure 11 illustrates the high-level control flow of the main application. The workflow initializes the session state. Then, the workflow renders the sidebar for chat history access and controls. After that, the workflow waits for user input. Upon receiving a new message, the app performs semantic search.

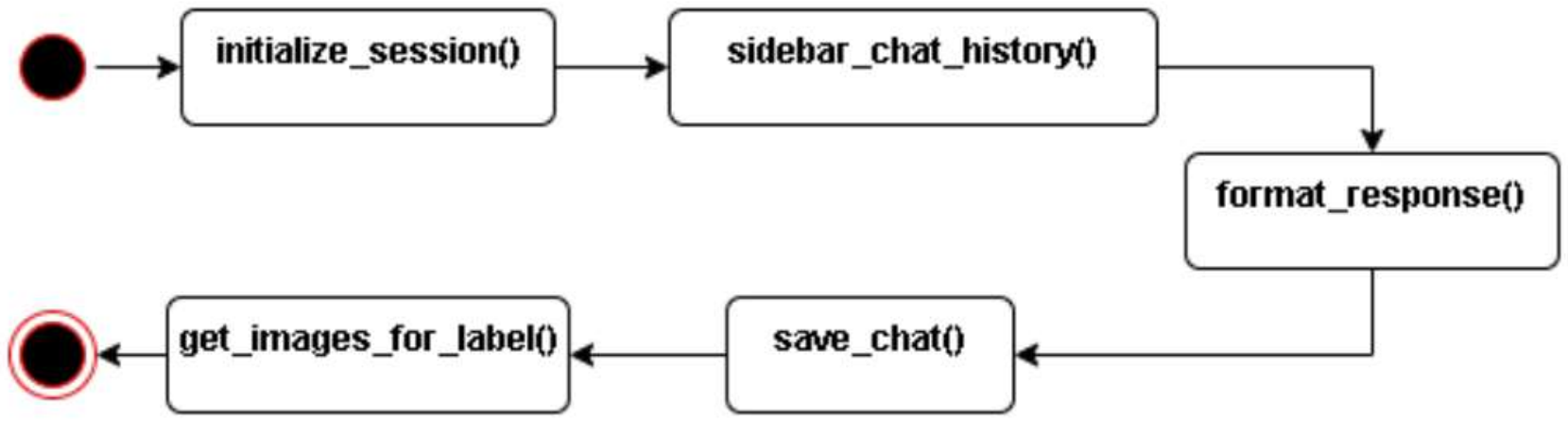

**Figure 11.** Logical flow of main application functions

Moreover, it also invokes the LLM in a prompt template. Afterwards, the next step identifies disease labels from the response and returns relevant images. All users and assistant messages are stored in a JSON-based chat history for later.

# Experimentation

## Chatbot Design and Operation

*The chatbot interface is built with Streamlit.* It supports natural language input and rich multimedia output to aid farmers and gardeners to seek plant health diagnostics (see Figure 12).

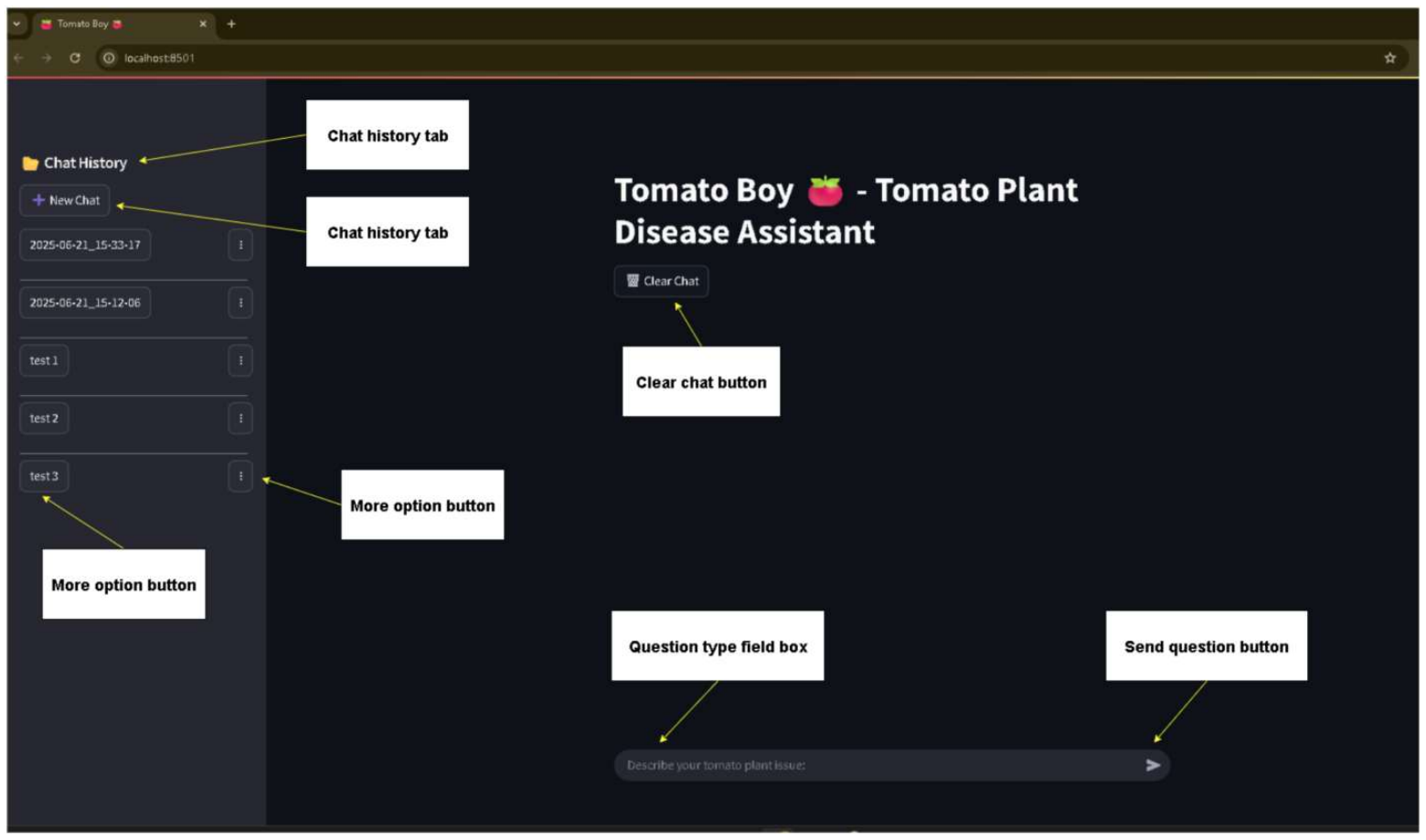

**Figure 12.** Main user interface of the chatbot

*Question Input and LLM Response Display.* Users interact with the chatbot by entering their queries into a text input box. As shown in Figure 13, upon submission, the question is processed through the LangChain pipeline, and a textual diagnosis is generated using the Mistral or LLama 3.2 model.

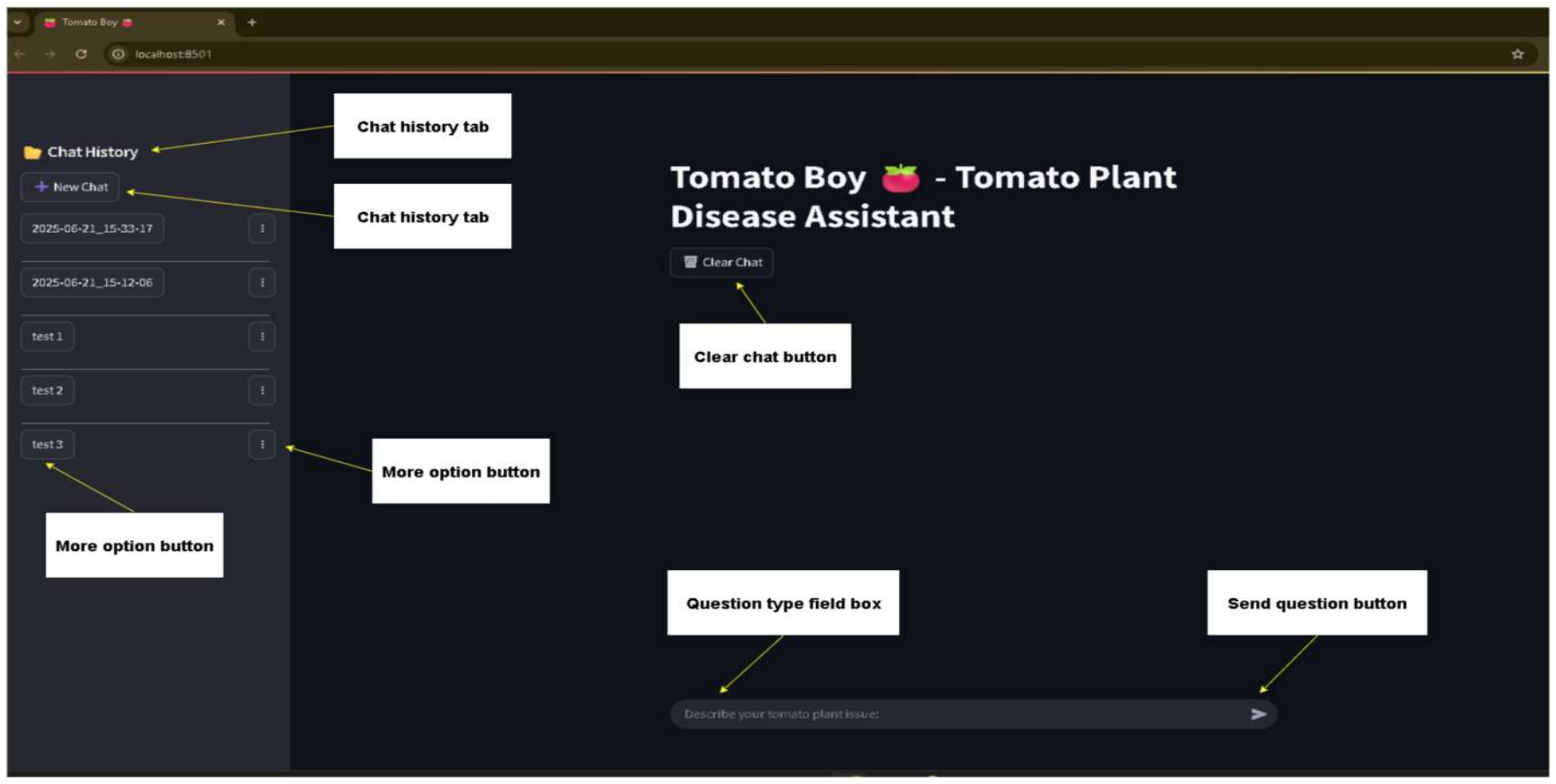

**Figure 13.** Text-based diagnostic response generated by the LLM

*Visual Feedback via Disease Image Matching.* In addition to textual output, the system enriches the response with diagnostic imagery if disease labels are identified in the LLM output. This is achieved through a keyword-matching mechanism followed by vector-based fallback retrieval. Relevant images are loaded and displayed as shown in Figure 14.

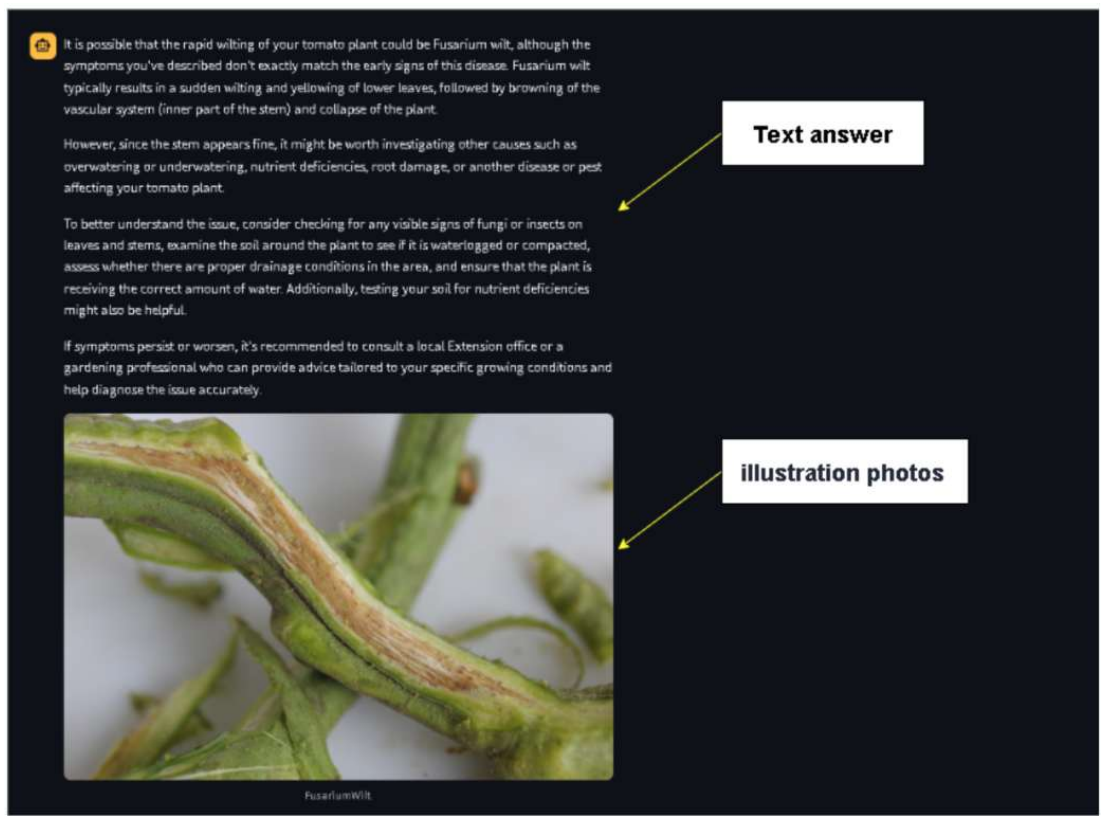

**Figure 14.** Illustrative disease image displayed based on label recognition

## Evaluation Metrics

Each model was evaluated based on the following metrics:

- *Response Time (s):* Time taken to generate a full response.
- *Answer Length (Tokens):* Approximate number of tokens in the generated output.
- *Relevance Score:* measures the semantic similarity between the chatbot response and the expected output using SentenceBERT embeddings.

$$Relevance\ Score = \cos(E_r, E_e) \qquad (1)$$

  where $E_r$ is the embedding vector of the generated response, $E_e$ is the embedding vector of the expected output, cos (□) denotes cosine similarity.

  The score ranges from 0 to 1, where higher values indicate greater semantic similarity.

- *Grammar Score:* evaluates the fluency and grammatical correctness of the generated response using LanguageTool**.**

$$Grammar\ Score = 1 - \frac{N_e}{N_w} \quad (2)$$

where Ne is the number of grammatical errors detected by LanguageTool, Nw is the total number of words in the response.

A higher score indicates better grammatical quality and fluency.

- *Memory Usage (MB):* Change in memory consumption during inference**.**

**Table 1. List of 5 query testcases**

| Query | Description |
|---|---|
| Q1 | My tomato leaves are curling and turning yellow, what's wrong? |
| Q2 | There are white spots under the leaves and a web-like structure. |
| Q3 | Fruits are rotting at the bottom before they ripen. |
| Q4 | My plant leaves have tiny holes, and some bugs are visible. |
| Q5 | The stems are wilting even though I water regularly. |

## Result Findings

We organize the list of 5 queries at increasing difficult level, that is, Q1 to Q5 which are described in Table 1. The reference answers are demonstrated in Table 2.

**Table 2. List of associated reference answers**

| Answer | Description |
|---|---|
| A1 | Yellow curling leaves may be caused by nutrient imbalance, pests like aphids, or viral infections. |
| A2 | White spots and webbing may indicate spider mite infestation. |
| A3 | Bottom rot is likely caused by blossom end rot due to calcium deficiency. |
| A4 | Tiny holes and visible bugs suggest flea beetles or cutworms. |
| A5 | Stem wilting despite watering may indicate bacterial or fungal wilt. |

Table 3 presents the evaluation results of the Mistral model in the chatbot. The response time ranges from 7.19 to 14.46 seconds which depend on the query complexity. The relevance scores are moderate between 0.37 and 0.64. So, the chatbot provides partially relevant answers to tomato disease questions. Meanwhile, the grammar scores remain very high from 0.99 to 1.00. Thus, the generated responses are grammatically correct and fluent. Memory consumption also remains stable during the experiments.

**Table 3. Mistral used chatbot**

| Query | Time (s) | Tokens | Relevance | Grammar | Memory (MB) |
|---|---|---|---|---|---|
| Q1 | 14.46 | 460 | 0.64 | 0.99 | 0.17 |
| Q2 | 8.55 | 430 | 0.63 | 0.99 | 0.00 |
| Q3 | 9.77 | 280 | 0.40 | 1.00 | -0.30 |
| Q4 | 8.75 | 290 | 0.37 | 0.99 | -0.26 |
| Q5 | 7.19 | 420 | 0.37 | 0.99 | 0.00 |

Table 4 shows the performance of Llama 3.2 model in the chatbot. Compared with Mistral, Llama 3.2 generally achieves higher relevance scores from 0.60 to 0.68 for most queries. The grammar quality is high with scores close to 1.00. The response time varies between 4.65 and 18.95 seconds. Overall, Llama 3.2 produces more semantically relevant responses. Whereas, both models maintain good grammatical quality and stable memory usage.

Comparing the both models, we can see that Mistral responds faster in complex queries (e.g., Q5). LLama 3.2 responds faster on simpler prompts like Q2. Moreover, LLama 3.2 achieved higher semantic relevance scores. It means that LLama indicates better contextual interpretation of queries. Remarkably, both models maintain near perfect grammar scores where LLama 3.2 scores slightly higher overall. While Mistral showed slightly more stable memory cost across the questions, LLama 3.2 had more variability. Besides, LLama 3.2 produces more verbose outputs. It shows that Llama 3.2 is helpful with clarity, however it is possible to increase latency.

Both models have strong performance. Mistral is better suited for environments which require faster and lightweight responses. In contrast, LLaMA 3.2 excels in semantic understanding and output fluency. It is preferable for high-accuracy use cases.

**Table 4. Llama 3.2 used chatbot**

| Query | Time (s) | Tokens | Relevance | Grammar | Memory (MB) |
|---|---|---|---|---|---|
| Q1 | 18.95 | 520 | 0.68 | 1.00 | 0.45 |
| Q2 | 4.65 | 450 | 0.65 | 0.99 | 0.00 |
| Q3 | 6.99 | 310 | 0.60 | 1.00 | -0.30 |
| Q4 | 8.63 | 305 | 0.66 | 0.99 | -0.24 |
| Q5 | 10.74 | 500 | 0.64 | 0.99 | -0.07 |

## Discussion and Conclusion

This paper presented an intelligent chatbot system that uses natural language interactions to help diagnose tomato plant diseases. The system provides users with fast, interpretable, and offline-capable support by integrating Streamlit as a lightweight web interface, LangChain as the coordination framework, ChromaDB for vector-based document retrieval, and Ollama to run local LLMs like Mistral and LLaMA 3.2. Overall, the implementation shows how open-source LLMs, vector search, and RAG techniques can be combined to solve domain-specific agricultural problems using easily accessible technology.

For the future works, as the current system supports 45 tomato diseases, we will expand the types of diseases. To assist more detailed result and a higher accuracy, we will expand the text and image datasets. Moreover, we will implement more complex multi modal to compare and select the best one with the best performance.